\documentclass{article}
\usepackage{ijcai20}

\usepackage{times}
\usepackage{soul}
\usepackage{url}
\usepackage[hidelinks]{hyperref}
\usepackage[utf8]{inputenc}
\usepackage[small]{caption}

\usepackage{array}
\usepackage{comment}
\usepackage{todonotes}
\usepackage[linesnumbered,ruled,vlined]{algorithm2e}
\usepackage{color,soul}
\usepackage{ctable}
\usepackage{pifont}

\usepackage{graphicx}
\usepackage{amsmath}
\usepackage{amsthm}
\usepackage{amssymb}

\usepackage{booktabs}

\usepackage{subcaption}

\newcolumntype{L}[1]{>{\raggedright\let\newline\\\arraybackslash\hspace{0pt}}m{#1}}
\newcolumntype{C}[1]{>{\centering\let\newline\\\arraybackslash\hspace{0pt}}m{#1}}

\newcommand{\bs}{\boldsymbol}

\SetKwComment{Comment}{$\triangleright$\ }{}

\begin{document}

\title{A Deep Ensemble Multi-Agent Reinforcement Learning Approach\\ for Air Traffic Control}

\author{
Supriyo Ghosh\footnote{Contact Author}$^\dag$\and
Sean Laguna$^\dag$\and
Shiau Hong Lim$^\dag$ \and
Laura Wynter$^\dag$ \And
Hasan Poonawala$^\ddag$ \\
\affiliations
$^\dag$IBM Research AI, Singapore 018983 \\
$^\ddag$ Amazon Web Services
\emails
supriyog@ibm.com,
\{slaguna, shonglim, lwynter\}@sg.ibm.com, hasanp1987@gmail.com
}

\maketitle

\begin{abstract}
Air traffic control is an example of a highly challenging operational problem that is readily amenable to human expertise augmentation via decision support technologies. In this paper, we propose a new intelligent decision making framework that leverages multi-agent reinforcement learning (MARL)  to dynamically suggest adjustments of aircraft speeds in real-time. The goal of the system is to enhance the ability of an air traffic controller to provide effective guidance to aircraft  to avoid air traffic congestion, near-miss situations, and to improve arrival timeliness. 
We develop a novel deep ensemble MARL method that can concisely capture the complexity of the air traffic control problem by learning to efficiently arbitrate between the decisions of a local kernel-based RL model and a wider-reaching deep MARL model. The proposed method is trained and evaluated on an open-source air traffic management simulator developed by Eurocontrol. 
Extensive empirical results on a real-world dataset including thousands of aircraft demonstrate the feasibility of using multi-agent RL for the problem of en-route air traffic control and show that our proposed deep ensemble MARL method significantly outperforms three state-of-the-art benchmark approaches.
\end{abstract}

\section{Introduction}
In 2018, the world's airlines served 4.3 billion passengers over roughly 22,000 routes, an increase of 1,300 routes and more than 200 million journeys from the previous year alone. This growth is expected to continue, with global passenger demand doubling in the next two decades \cite{iata2019feb}.
 
With the increase in passenger demand for air travel as well as new forms of urban aircraft, from air taxis to drones, the densification of air traffic will require more sophisticated technologies, including automation and decision support. 
Indeed, air traffic controllers  will have difficulty handling the volume increases predicted for the next decade and beyond.
Industry experts are thus calling for and embarking on the development of intelligent  assistants to support pilots and controllers with real-time decision-making \cite{thales}.

The  challenges of  automating air traffic control (ATC) include dealing with multiple decision-makers in a highly dynamic and stochastic environment. Decisions taken by one aircraft  impact the outcome of that  and other aircraft in the future. The problem is thus inherently sequential, and involves multiple actors influencing each other. An aircraft must consider local information such as  other aircraft flying in its sector, and global information such as congestion in other sectors. ATC bears some resemblance to operational control problems in autonomous driving \cite{autonomous} and in robotics  such as robot soccer \cite{soccer} and UAVs \cite{drones}. In those settings, multi-agent reinforcement learning (MARL) methods have been successfully applied, albeit to problems with relatively small numbers of agents.
 
We  address the problem of learning cooperative policies in this complex, partially observable multi-agent domain  through MARL techniques. We propose a novel general purpose deep ensemble MARL method that can effectively capture the multi-agent interactions during training and efficiently learn to arbitrate between the decisions of multiple pre-trained policies during execution. We further demonstrate that the complexity of the ATC problem can be concisely  captured by learning an optimized ensemble of two pre-trained policies: (i) a kernel-based approach employing an agent-centric state representation that captures information in the neighborhood of each aircraft; and (ii) a deep MARL model that takes into consideration additional  wider-reaching state information. In order to be scalable, our approach is fully decentralized where each aircraft is considered an agent.

Kernel-based methods perform well if a state encountered during testing has a dense neighborhood of training examples, but  extrapolate poorly. In contrast, a deep MARL method is more flexible and able to generalize well, but is prone to various pathologies and can be brittle, even in regions of dense training data. Our proposed deep ensemble MARL method combines their strengths by efficiently learning when to switch between the two policies during execution. 
To that end, our key contributions are as follows:
\begin{itemize}
\item We propose to model the air traffic control problem using a MARL framework which, different from existing approaches, seeks to minimize the combined cumulative long-term effect of conflicts, congestion, delays and fuel costs.
\item We develop a kernel-based approach employing agent-centric local information to represent the aircraft's state.
\item We  design a deep MARL model offering an extended view with richer state information. 
\item Using these, we develop a novel ensemble learner that leverages the pre-trained policies from the kernel and deep MARL techniques to efficiently arbitrate the complex boundary between the effectiveness of these two methods.
\item We conduct extensive experiments on a large real-world data set consisting of thousands of aircraft and empirically demonstrate that our proposed ensemble learner significantly outperforms  state-of-the-art benchmarks: (a) a baseline approach, (b) a local search approach, and (c) the MARL model of \citeauthor{brittain2019autonomous} \shortcite{brittain2019autonomous}.
\end{itemize}

 \section{Related Work}
 The automation of air traffic control was addressed more than a decade ago in 
\citeauthor{wollkind2004automated} \shortcite{wollkind2004automated}, who provide an incremental multi-agent bargaining process between two   aircraft  so as to choose a solution that is Pareto optimal. 
In \citeauthor{farley2007fast} \shortcite{farley2007fast}, the authors propose an auto-resolver that iteratively computes candidate air traffic trajectories until a suitable trajectory is found that satisfies all of the conflict resolution conditions. Closer to our setting, \citeauthor{tumer2007distributed} \shortcite{tumer2007distributed} design a comprehensive reward function within a reinforcement learning approach, but consider each agent as a fixed geographical location independent of other agents, due to computational limitations. Their learning process did not take into account state transition dynamics. 
Given improvements in computer hardware and deep learning over the past decade, we are able to propose a  more efficient and effective method  than the early approaches cited above. We aim  to compute real-time cooperative multi-agent policies with a realistic set of ATC features, namely potential conflict avoidance, congestion reduction, fuel costs and arrival timeliness.
 
The  problem of ATC automation using MARL was  addressed more recently by \citeauthor{brittain2019autonomous} \shortcite{brittain2019autonomous}. 
However, in that work, the authors limited their model to handling only potential aircraft conflict resolution, which reduces significantly the complexity of their model. As discussed  in \citeauthor{tumer2007distributed} \shortcite{tumer2007distributed}, air traffic congestion and arrival timeliness are important factors, and conflict resolution in the absence of those considerations will typically lead to unrealistic solutions such as excessive arrival delays. Indeed, it is relatively straightforward to devise policies whereby aircraft conflicts are avoided whilst increasing  congestion, causing arrival delays, or excessive fuel costs. 

Our work advances the state of the art in ATC automation through  the development of  a   cooperative MARL framework which leverages a compact reward function that includes penalties for conflict, congestion, arrival delays and fuel cost and does so in a scalable manner that can handle wide-area flight zones with thousands of aircraft. In addition, we show that our method significantly outperforms that of \citeauthor{brittain2019autonomous} \shortcite{brittain2019autonomous}.

There is a substantial literature on cooperative MARL in general, whereby multiple agents jointly learn a policy to optimize a cumulative future expected reward \cite{bu2008comprehensive}, summarized well in the recent survey paper by \citeauthor{hernandez2019survey} \shortcite{hernandez2019survey}. Solving a cooperative MARL problem as a centralized single-RL problem is intractable since the complexity of the problem increases exponentially with the number of agents due to the joint action or state space. While a straightforward workaround is to consider the agents independently, such as through independent Q-learning \cite{tan1993multi}, that leads generally to unstable learning and non-stationarity \cite{matignon2012independent}.
A popular approach is to perform centralized learning with decentralized execution \cite{lowe2017}. Our approach  corresponds in a similar vein to learning a decentralized, shared policy among homogeneous agents, as in \citeauthor{gupta2017} \shortcite{gupta2017}.

We propose a novel ensemble approach to this complex MARL problem. While there are several works that have proposed ensemble methods for  (single-agent) reinforcement learning (RL), we are unaware of such in the MARL domain. Existing methods for single-agent ensemble RL rely on either (a) averaging the Q-values from multiple target networks so as to improve stability during the learning process \cite{anschel2017averaged,chen2018ensemble} or (b) performing policy-level aggregation during testing through mechanisms such as majority voting or Boltzmann multiplication \cite{wiering2008ensemble,hans2010ensembles}. The former  cannot, however, take advantage of kernel-based methods, and the latter is not suitable when there are multi-agent interactions. Our proposed approach thus involves  a separate master deep ensemble learner to handle the multi-agent interactions during training so as to efficiently arbitrate between these two policies.

In a similar spirit to our approach, but in a single-agent RL setting, \citeauthor{peng2018improving} \shortcite{peng2018improving} designed an ensemble method for personalized medical recommendations  that learns  probabilistic weights  for a deep RL and a kernel policy.  In that work, the discounted return for an ensemble policy can be computed using off-policy evaluation with a fixed set of  patient treatment trajectory data. In our multi-agent setting, however, their method would not work, as switching  policies across  agents according to a probabilistic weight would drastically deteriorate performance. In contrast, our ensemble method involves training a separate master deep neural network to handle the effects of multi-agent interactions during training by taking advantage of a real-world simulator.

\section{Methodology}
A standard reinforcement learning (RL) problem is typically defined with respect to a Markov Decision Process (MDP),
given by a tuple $\langle S, A, P, R, \gamma \rangle$, where $S$ denotes the set of all possible states in the environment, $A$ represents the set of permissible actions for the agent, $P: S\times A\rightarrow \mathcal{P}(S)$ denotes
the transition function providing the next-state distribution after executing action $a\in A$ in state $s\in S$,
$R: S\times A \rightarrow \mathbb{R}$ denotes a reward function providing the expected immediate reward for executing $a$ in $s$, and $\gamma \in [0,1]$ is a discount factor.

A (potentially randomized) policy $\pi: S \to\mathcal{P}(A)$ specifies the action to take in each state $s\in S$. The $Q$-value
of a state-action pair $(s,a)$ with respect to a policy $\pi$ is given by the expected total reward after executing $a$ in $s$
and following $\pi$,
\[ Q^\pi(s,a)=\mathbb{E}_\pi\left[\sum_{t=0}^\infty \gamma^t R(s_t,a_t) | s_0=s \right]. \]
The goal of the agent is to learn a policy $\pi^*$ to maximize $Q(s,a)$ for all $(s,a)$.
In finite-state MDP,  algorithms such as tabular Q-learning
\cite{watkins1992q} can be shown to asymptotically learn the optimal $Q$ function. For large or continuous state spaces,
some form of function approximation is typically needed. The design of state features  plays an important role.

\subsection{Kernel Based RL}
Kernel Based RL (KBRL) \cite{ormoneit2002kernel} is one such Q-value approximation scheme which has the benefit of having strong theoretical properties. 
In particular, KBRL algorithms can be proven to converge to a unique fixed point and the approximation error can be bounded asymptotically. KBRL is a model-based approach relying on a set of sample transitions $\mathbb{S}\equiv \{s^a_k, r^a_k,\hat{s}^a_k | k=1,2,...,n_a\}$ associated with each action $a\in A$. The complexity of KBRL increases with the number of sample transitions and the size of the state space, thereby limiting its applicability in practice.
KBRL was thus extended to achieve better scalability through a modification known as Kernel Based Stochastic Factorization (KBSF) \cite{barreto2016practical}. In contrast to KBRL wherein each sample transition gives rise to a state, KBSF generates a set of representative states and uses stochastic factorization to reduce the size of the transition matrix.

\begin{algorithm}[!htb]
	\textbf{Inputs:} $S^a = \{s^a_k, r^a_k, \hat{s}^a_k | k = 1, 2, ..., n_a\} \forall a\in {\mathcal A}$\;
	$\bar{S} = \{\bar{s}_1, ..., \bar{s}_m\}$ \Comment{\small $m$ representative states}
	\textbf{Outputs:} $Q^*$ matrix of size $|\bar{S}| \times {\mathcal A}$\;
	Define Gaussian kernel $\bar{\kappa}_{\bar{\tau}}$ and $\kappa^a_{\tau}$ using Equation~\ref{eq:gkernel}\;
	\For{each $a\in {\mathcal A}$}{
		Compute matrix $D^a : d^a_{ij} = \bar{\kappa}_{\bar{\tau}}(\hat{s}^a_i, \bar{s}_j)$ \;
		Compute matrix $K^a : k^a_{ij} = \kappa^a_{\tau}(\bar{s}_i, s^a_j)$ \;
		Compute reward $r^a : r^a_i = \sum_j k^a_{ij} r^a_j $\;
		Compute transition probabilities: $P^a = K^a D^a$ \;
	}
	Solve MDP $\{ \bar{S}, {\mathcal A}, P^a, r^a, \gamma = .99\}$ and obtain $Q^*$\;
	\Return $Q^*$
	\caption{\sc{SolveKBSF}()}
	\label{algo1:KBSF}
\end{algorithm}

The KBSF algorithm is outlined in Algorithm~\ref{algo1:KBSF}. For the KBSF method, we first collect $n$ transitions $(s_t, a_t, r_t, s_{t+1})$ by simulating a random policy. We then compute $m$ representative states, $\bar{S} = \{\bar{s}_1, ..., \bar{s}_m\}$ from $n$ sample transitions using \textit{k-means} clustering. Next, we define a Gaussian kernel, $\bar{\kappa}$ to build the components of the transition matrix of the MDP:
\begin{flalign}
\hspace{-0.1in}\bar{k}_{\bar{\tau}}(s,s') = \bar{\phi}\left(\frac{|| s-s' ||}{\bar{\tau}}\right); \bar{\kappa}_{\bar{\tau}}(s,\bar{s}_i) = \frac{\bar{k}_{\bar{\tau}} (s,\bar{s}_i)}{\sum_{j=1}^m \bar{k}_{\bar{\tau}} (s,\bar{s}_j)} \label{eq:gkernel}
\end{flalign}
where $\mathcal{k} . \mathcal{k}$ measures the distance under a Euclidean norm, $\bar{\phi}(x)=\exp(-x^2)$,
and $\bar{\tau}$ denotes the kernel width. Using the Gaussian kernel from Equation~\eqref{eq:gkernel}, we define: (a) $D^a$ a matrix of size $n\times m$ with kernel width $\bar{\tau}$; $D^a : d^a_{ij} = \bar{\kappa}_{\bar{\tau}}(\hat{s}^a_i, \bar{s}_j)$ and (b) $K^a$ a matrix of size $m\times n$ with kernel width $\tau$; $K^a : k^a_{ij} = \kappa^a_{\tau}(\bar{s}_i, s^a_j)$. The reward function is computed as $r^a : r^a_i = \sum_j k^a_{ij} r^a_j $ and the transition probability matrix is defined as $P^a = K^a D^a$. 
The MDP is solved by \textit{Policy Iteration} \cite{howard1960dynamic} to obtain the $Q^*$ values for the representative state-action pairs. Finally, during the validation phase, we compute the policy $\bar{\pi}(s)$ for any given state $s$ through 
 \begin{flalign}
\bar{Q}(s,\!a) \!=\!\! \sum_{i=1}^m \!\bar{\kappa}_{\bar{\tau}}(s, \!\bar{s}_i)Q^*(\bar{s}_i, \!a); \bar{\pi}(s)\!=\! \arg\!\max_a \bar{Q}(s,\!a) .\label{eq:kernelTest}
\end{flalign}

\subsection{Deep MARL}
To solve the Q-learning problem with a much larger state-space, \citeauthor{mnih2015human} \shortcite{mnih2015human} proposed using a neural network approximator which they termed deep Q-learning, or DQN.
DQN uses a parameter $\theta$ learnt using gradient descent on the loss function:
${\mathcal L}_{\theta} = \mathbb{E} \big[ (y^{DQN} - Q(s,a;\theta))^2\big ]$, 
where $y^{DQN} = r+\gamma \max_{a'}Q(s',a';\theta^-)$ is the target value. 

Numerous methods have since then been proposed for RL with neural network approximators, including those based on policy gradient \cite{mnih2016asynchronous}. The Proximal Policy Optimization (PPO) \cite{schulmanWDRK17} in particular was introduced to overcome the shortcomings of approaches based on policy gradient. 
PPO has been shown to offer better performance than both DQN and advantage actor-critic (A2C) \cite{mnih2016asynchronous}. PPO as its name implies, seeks to find a proximal policy, and as such to avoid large policy updates. Let $\theta$ denote the parameters of the policy network at time $t$ and $r_t(\theta)$ represent the ratio $\frac{\pi_{\theta}(a_t|s_t)}{\pi_{\theta_{old}}(a_t|s_t)}$. The PPO loss function is given by equation \eqref{eq:ppoClip}, where $\epsilon$ is a hyperparameter that determines the bound for $r_t(\theta)$ and $A_t := R_t - V(s_t)$ is an estimator of the advantage function at time $t$.
\begin{flalign}
\hspace{-0.08in}{\mathcal L}^{{\small CLIP}}\!(\theta) \!=\! \mathbb{E}_t\big [\! \min(r_t(\theta)A_t, clip(r_t(\theta),\!1\!-\!\epsilon,\!1\!+\!\epsilon) A_t )\big ] \label{eq:ppoClip}
\end{flalign}

Analogous to \citeauthor{gupta2017} \shortcite{gupta2017}, we propose learning a decentralized, shared policy among homogeneous agents to solve this cooperative multi-agent decision problem.
This approach can be seen as solving a multi-agent cooperative Markov game, where all agents share and update the same policy (i.e., $\bs{\pi}_{\theta}$) simultaneously. The game is defined by $N$ agents with their corresponding actions ${\mathcal A}_1, {\mathcal A}_2, ..., {\mathcal A}_N$. Let ${\mathcal S}$ denote the global state space and ${\mathcal O}_1, {\mathcal O}_2, ..., {\mathcal O}_N$ represent  local observations of the agents. In each time step, the agents execute their actions based on the current policy and local observations  producing the next state depending on the transition function ${\mathcal T}: {\mathcal S} \times {\mathcal A}_1 \times {\mathcal A}_2 \times ... \times {\mathcal A}_N \rightarrow {\mathcal S}$. After the transition, each agent $i$ receives a local observation $o_i: {\mathcal S} \rightarrow {\mathcal O}_i$ and a reward $r_i: {\mathcal O}_i \times {\mathcal A}_i \rightarrow \mathbb{R}$.
We follow this approach in training the kernel-based and deep MARL policies.

We begin the training process by initializing the policy parameters $\theta$ to $\theta_0$. 
In each episode $k$, the environment is reset to an initial state $s_0$. Let $N_t$ denote the number of active agents at time $t$. In each time step $t$, all active agents sample an action using the current policy parameters $\theta_k$. The joint action $\bs{a}_t = (a_t^1,...,a_t^{N_t})$ is then executed and the immediate rewards are observed along with the subsequent state. The $N_t$ sample transitions of the current episode are also stored in a buffer $D$. After accumulating the samples for the current episode, $M$ rounds of update are performed with a minibatch of transitions sampled from buffer $D$ using stochastic gradient descent (SGD) on the loss function of equation \eqref{eq:ppoClip}.

\subsection{Deep Ensemble MARL}
\begin{figure}[!htb]
	\centering
	\includegraphics[width=0.47\textwidth]{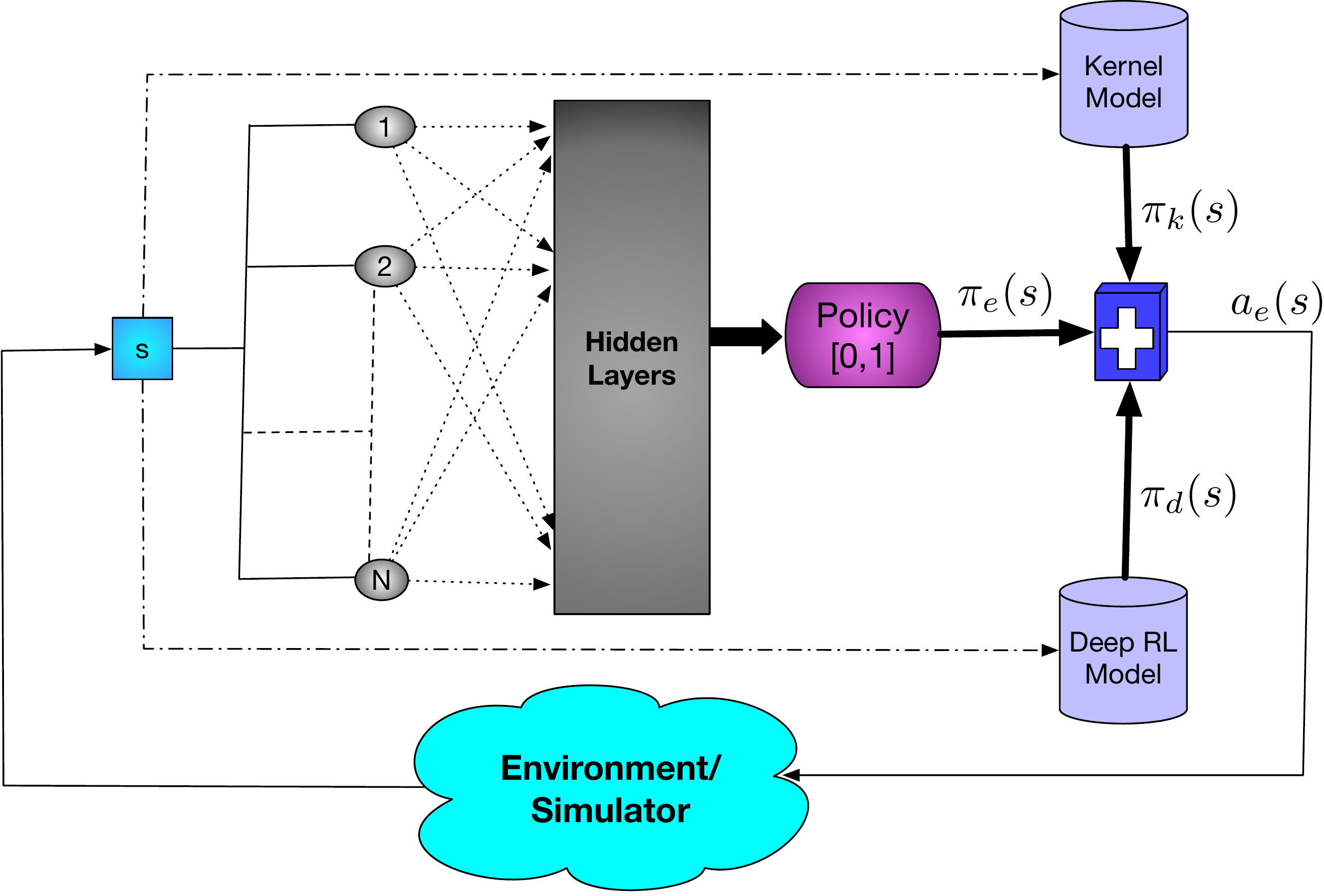}
	\caption{ Architecture of the ensemble approach.}
	\label{fig:ensembleArchitecture}
\end{figure}

The KBSF has the advantage of its theoretical convergence and asymptotic bounds. We expect KBSF to perform very well in state regions with dense training examples. However, this is only feasible if the dimension of the state-space
remains small. We can therefore  afford a relatively compact state representation that includes only a local view
of the global state space.
With a deep MARL policy we can afford to employ a more ambitious state representation. Training a deep RL policy, however, can be challenging.  Specifically, the resulting policy may have pathologies due to non-linear function approximation and can be brittle in certain state regions. 
We therefore propose an ensemble learner that is able to combine these two approaches across the highly complex boundary between them. 
We develop a novel ensemble learning approach wherein we train a separate deep neural network to leverage the pre-trained KBSF and Deep MARL policies to obtain our final  ensemble policy.
The action space consists of two actions:  choose the KBSF policy or choose that of the deep MARL, and we employ the same set of features used in the KBSF method to represent the state space. The architecture of our deep ensemble method is shown in Figure~\ref{fig:ensembleArchitecture}. 

The steps of our proposed deep ensemble MARL are provided in Algorithm \eqref{algo1:ensembleRL}. We begin by loading the trained kernel model $\tilde{K}$ and the learned deep MARL policy $\tilde{\pi}$ parameterized by $\tilde{\theta}$. Let $\theta$ denote the parameters of the policy network of our ensemble MARL initialized to $\theta_0$. In each episode $k$, we reset the environment to an initial state $s_0$ with a new agent scenario. At each time step $t$, all active agents sample an action using the current ensemble policy parameters $\theta_k$. If the sampled action $a^i_t$ for agent $i$ is 0, then we choose an action $\tilde{a}^i_t$ recommended by the KBSF policy and otherwise, the action $\tilde{a}^i_t$ is sampled using the deep MARL policy $\tilde{\pi}(\tilde{\theta})$. 
We execute the joint action $\bs{\tilde{a}}_t = (\tilde{a}_t^1,...,\tilde{a}_t^{N_t})$ in the environment and the agents obtain the immediate rewards along with the data from the next state. The $N_t$ transitions $(s_t^i, a_t^i, r_t^i, s_{t+1}^i)$ are stored in a buffer $D$. Once all the samples are generated for the current episode, we perform $M$ rounds of update to the current policy parameters $\theta_k$ with a minibatch of transitions sampled from buffer $D$ to optimize the surrogate loss function \eqref{eq:ppoClip}.

\begin{algorithm}[!htb]
	\emph{Initialize} policy parameters to $\theta_0$\;
	\emph{Load} learned kernel model $\tilde{K}=(\tilde{S}, \tilde{Q}, \tilde{\kappa_{\tau}}, \tilde{\mu}, \tilde{\sigma})$\;
	\emph{Load} learned deep MARL model and policy $\tilde{\pi}(\tilde{\theta})$\;
	\For{$k = 1$ to max\_iterations}{
		\emph{Initialize} an empty buffer $D$ to capacity $C$\;
		Reset environment to state $s_0$ with a schedule $P_k$\;
		\For{$t=1$ to $T$}{
			\For{$i=1$ to $N_t$}{
				Sample action $a_t^i$ according to current observation $s_t^i$ and policy $\pi(\theta_k)$\;
				\If{$a_t^i == 0$}{
					Compute action $\tilde{a}^i_t$ using kernel parameters $\tilde{K}$ in equation \eqref{eq:kernelTest}\;
				}
				\Else{
					Sample action $\tilde{a}_t^i$ using learned deep RL policy $\tilde{\pi}(\tilde{\theta})$ for observation $s_t^i$\;
				}
			}
			Execute the joint action $\bs{\tilde{a}_t} = (\tilde{a}_t^1,...,\tilde{a}_t^{N_t})$ in simulator and get the next state $\bs{s}_{t+1}$\;
			\For{$i=1$ to $N_t$}{
				Store transition $(s_t^i, a_t^i, r_t^i, s_{t+1}^i)$ and the advantage estimator $A^i_t$ in $D$\;
			}
		}
		\For{$m=1$ to $M$}{
			Sample a minibatch of transitions from $D$ \;
			Update policy parameters $\theta_k$ to maximize the surrogate PPO objective function \eqref{eq:ppoClip}\;
		}		
	}
	\caption{\sc{SolveEnsembleMARL}()}
	\label{algo1:ensembleRL}
\end{algorithm}

\section{Experimental Design}
In this section, we describe the modeling aspects of our approach with respect to the ATC problem and the setup of our experiments as well as the main implementation issues concerning the simulator.

\subsection{Air Traffic Control Formulation} \label{sec:MARLfeatures}
The multi-agent model is defined such that  each aircraft is represented as an agent. 

\begin{figure*}[!htb]
	\centering
	\begin{subfigure}{0.3\textwidth}
		\includegraphics[width=\textwidth]{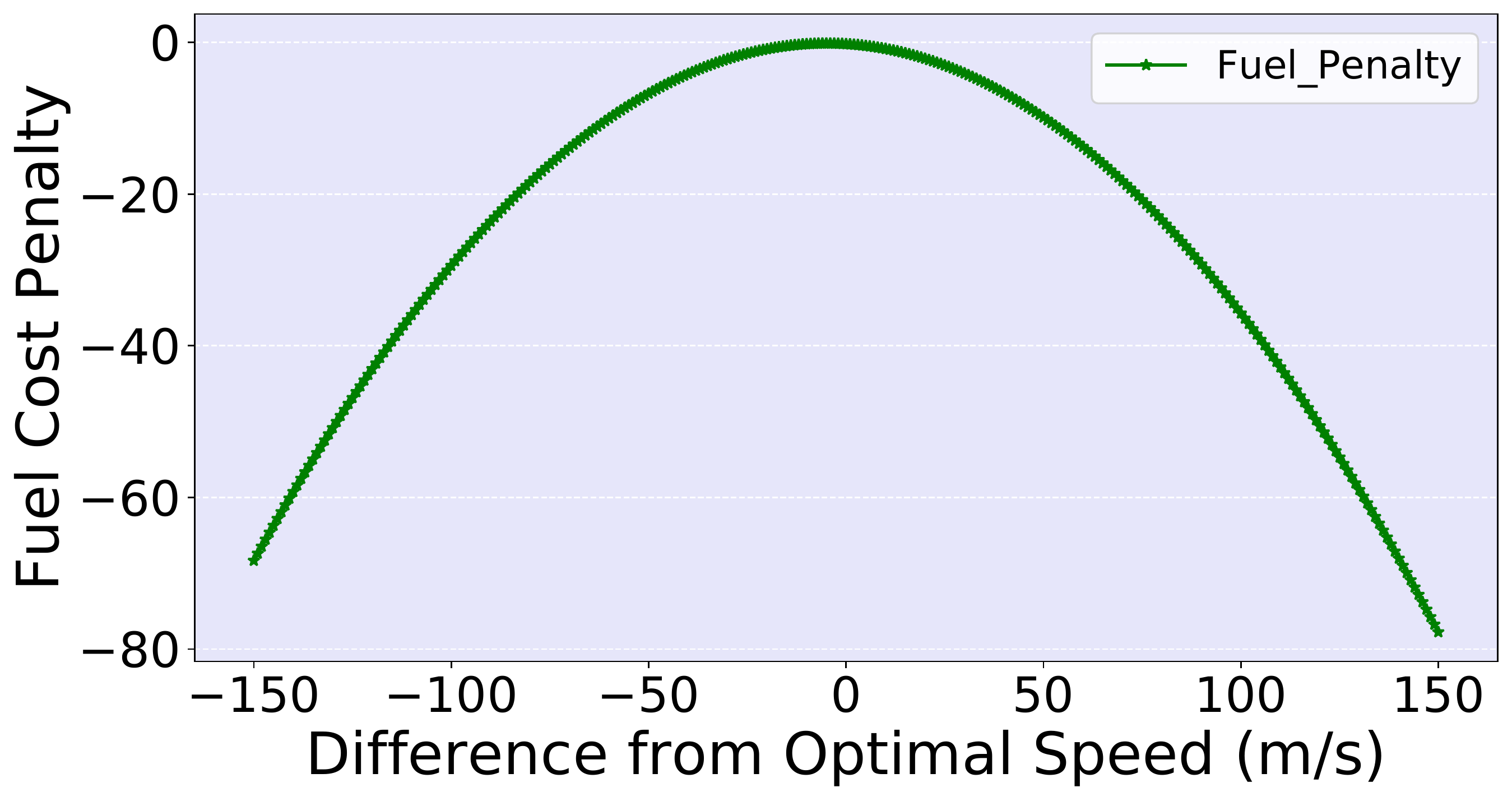} \caption{}
	\end{subfigure} 
	\begin{subfigure}{0.3\textwidth}
		\includegraphics[width=\textwidth]{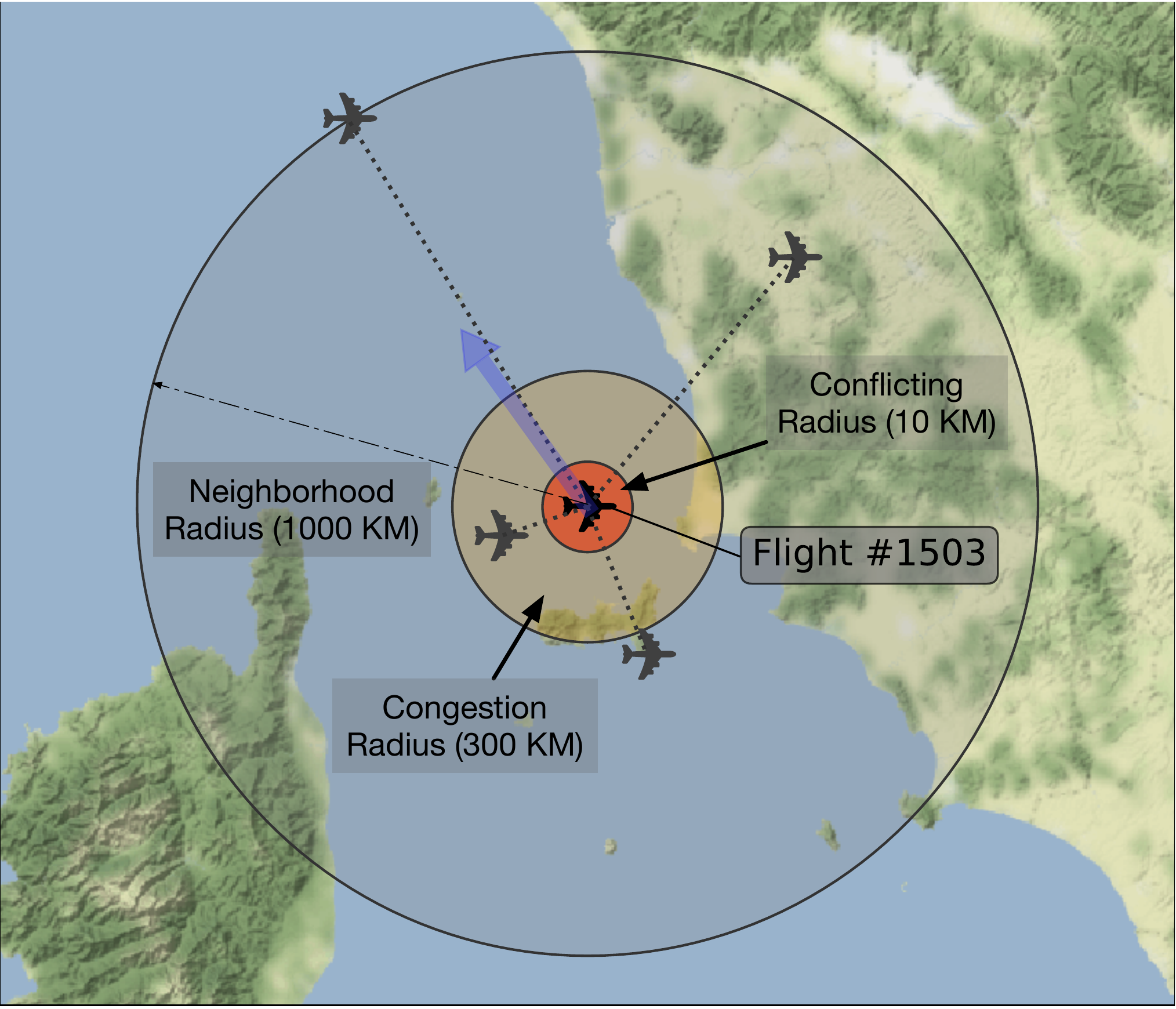} \caption{}
	\end{subfigure} 
	\begin{subfigure}{0.32\textwidth}
		\includegraphics[width=\textwidth]{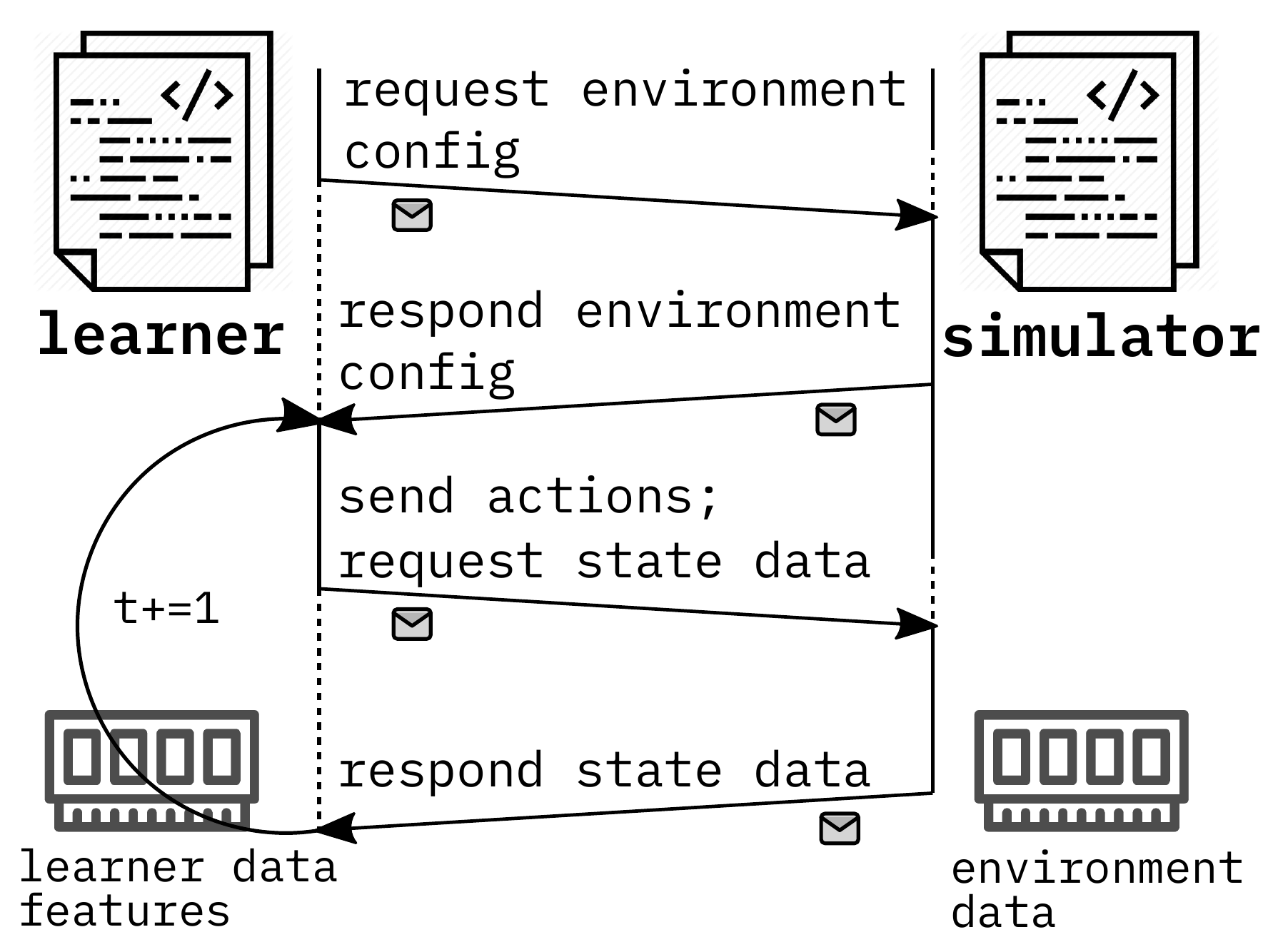} \caption{}
	\end{subfigure} 
	\vspace{-0.1in}
	\caption{(a) Fuel cost penalty function; (b) State representation used to model en-route ATC; (c) Message passing architecture built to interface between the ATC simulator and an external RL agent.}
	\label{fig:architecture}
	\vspace{-0.15in}
\end{figure*}

{\sc State space}: It is important for the efficacy of any RL framework to  define a rich yet compact representation. As such for the ATC problem we include the following information in the state:
\begin{itemize}
\item \textit{ Local information:} the aircraft's geographical location (i.e., latitude and longitude), velocity, direction, distance to the destination and timeliness of the journey with respect to the scheduled arrival time; and
\item \textit{ Neighboring aircraft  information:} For the $N$ nearest aircraft within a given radius, the relative distance and relative velocity of the aircraft. 
\end{itemize}
Figure~\ref{fig:architecture}(b)   illustrates  the state representation. Defined this way, the state space remains of constant size as the number of agents increases. 

{\sc Action space}: 
Our model allows for aircraft speed increase, decrease or no change. Specifically, at every time step, each agent decides whether to maintain, increase, or decrease the aircraft speed by $\delta$ meters/second, bounded by a minimum and maximum speed. That is, $$A_t = \{ \max(v_{min}, (v_{t-1} - \delta)),v_{t-1},\min(v_{max}, (v_{t-1} + \delta))\}.$$

{\sc Reward function}: The reward includes the following four components:
\begin{itemize}
\item \textit{Conflict cost:} To ensure a safe separation distance, if the relative distance between two aircraft is less than a threshold, a penalty cost is imposed;
\item \textit{Congestion cost:} To avoid congestion, if the number of aircraft within a given congestion radius exceeds a threshold, a penalty is imposed;
\item \textit{Lateness cost:} To minimize the delays, a penalty is imposed if the aircraft is behind schedule, where the penalty is proportional to the delay; and
\item \textit{Fuel cost:} The fuel cost penalty is defined as a quadratic function of the deviation of speed from the optimal speed for that type of aircraft.
\end{itemize}
Formally, the reward for an agent $i$ at time $t$ is:
\begin{small}
$$r^i_t \!=\! \alpha \cdot I(o_t^i,\! R^s)+\beta \cdot I(o_t^i,\! R^c,\! N^c)+\gamma \cdot \max(0, \tilde{d}^i_t - d^i_t)+\delta \cdot F(v_t^i - v_0^i)$$
\end{small}
The value of $I(o_t^i, R^s)$ is set to 1 if the relative distance from the nearest aircraft is less than the separation radius $R^s$ according to the current observation $o_t^i$, and otherwise it is set to 0. The value of $I(o_t^i, R^c, N^c)$ is set to 1 if the number of aircraft within the congestion radius $R^c$ exceeds the threshold value $N^c$, and 0 otherwise. Let $\tilde{d}^i_t$ denote the expected distance to travel by aircraft $i$ at time $t$ according to the given schedule and $d^i_t$ denote the actual distance traveled; then $\max(0, \tilde{d}^i_t - d^i_t)$ characterizes the amount of delay at time $t$. $F(v_t^i - v_0^i)$ represents the quadratic fuel cost function depending on the deviation of current speed from the optimal speed $v_0^i$, which is shown in Figure~\ref{fig:architecture}(a).
Finally, $\alpha, \beta, \gamma$ and $\delta$ denote the weights for conflict, congestion, delay and fuel cost penalties, respectively.

We extend the state space for the Deep MARL so as to enable a richer representation of the domain. For the deep MARL, the state space includes additional two sets of $3\times 3$ grid image information, a coarse grid that covers a large area and a fine grid that covers a smaller area around the aircraft. Each cell is identified by a Geo-hash encoding and the cell size, governed by the precision of the Geo-hash, contains the information about the number of aircraft present in that cell. 

\subsection{Air Traffic Simulator}
We employ an open source air traffic simulator developed by Eurocontrol for training and evaluation \cite{gurtner2017empirically}. The simulator was developed as part of the Enhanced Large Scale Aeronautical Telecommunication Network deployment (ELSA) consortium funded by the Single European Sky Air Traffic Management Research (SESAR) program\footnote{The open-source ELSA codes are accessed from https://github.com/ELSA-Project/ELSA-ABM}. The training and testing scenarios make use of aircraft schedule data from a region of airspace in southern Europe.

The ELSA Agent-Based Model (ELSA-ABM) simulator allows for simulating any number of aircraft within a set of adjacent sectors. The path of an aircraft $f$ is defined by a sequence of navigation points ${\mathcal P}^f=\{p^f_1, p^f_2, ...,p^f_n \}$ where each point $p^f_i \in {\mathcal P}^f$ is represented by a tuple $\langle l_i, t_i, s_i \rangle$. 
The location $l_i$ is specified by a latitude, longitude, and altitude value, $t_i$ represents the scheduled time to reach the navigation point and $s_i$ represents the expected speed between the point $p^f_i$ and $p^f_{i+1}$. At each time-step, the simulator updates the positions of an active aircraft using the \textit{Haversine} distance according to the current location, speed and other parameters (e.g., wind speed) that may affect the aircraft's movement.

\subsection{Message Passing Architecture}
Due to heavy computational requirements, the ELSA-ABM simulator is written in C and computes in-memory updates of all aircraft data. However, it does not provide an easy interface for an RL agent to interact with it as an environment. Therefore, we designed and built an adapter for the simulator to allow an external RL agent to extract the aircraft's state and to submit actions via a predefined set of definitions. To do so, we designed a two-way message passing system between the simulator and an RL agent using the Nanomsg Next Generation (NNG) library.
At each time step, the simulator collects all necessary state and reward information for all the active aircraft and sends it as serialized data to the agent who in turn sends the set of actions to execute to all the active aircrafts (refer to Figure \ref{fig:architecture}(c)). 

\subsection{Experimental Settings}
Our real-world  dataset includes flights from one 24-hour period and includes, for 1668 aircraft, the source, destination and departure time, a set of navigation points with their corresponding geographical locations and the expected arrival time and speed of each aircraft at each navigation point.
From this set, we generate 1000 training and 30 test scenarios by introducing random delay into the real-world data, ranging from -30 to +30 minutes around the aircraft's (true) departure time. 

The 24-hour period is subdivided into 360 time steps, each lasting 4 minutes. In the results presented here, we set the conflict, congestion and neighborhood radii  to 10, 300 and 1000 kilometers, respectively, and the penalties   to -1000 for each conflict, -100 for congestion. The delay penalty for an aircraft was set to -1 for every kilometer it is behind of its expected location. 
The fuel cost penalty follows the quadratic function illustrated in Figure~\ref{fig:architecture}(a), which is centered at the average expected speed for each aircraft. We consider three fuel cost scenarios: high, medium and low. Figure~\ref{fig:architecture}(a) illustrates the medium fuel cost scenario.

We compare our ensemble method with the following three benchmark approaches:

\noindent \textit{Baseline:} The flight information in our data set allows us to derive the expected speed between each pair of subsequent navigation points. We thus define the baseline solution by simulating the aircraft using these expected speeds.

\noindent \textit{Local search:} To test whether a local search method can outperform a reinforcement learning method that takes into account sequential decisions over a longer horizon, we implemented  a myopic best improvement-based local search method. At each time step, each active aircraft chooses its best action (the action with the highest global reward)  assuming that other agents will maintain their same speed from the previous time step.

\noindent \textit{DDMARL:} The third comparison is with the recently-published  deep distributed MARL
method of \citeauthor{brittain2019autonomous} \shortcite{brittain2019autonomous}.

\section{Empirical Results}
In this section, we provide  performance  results. The experiments were performed on an Ubuntu 16.04 virtual machine with 8-core CPU, 64 GB of RAM, and a single Nvidia Tesla P100 GPU. The distributed Ray framework and RLlib \cite{liang2017ray} were used for the PPO method.

\subsection{Training performance comparison}
For the training process, 600,000 transition samples were generated using a random policy for the KBSF method. These samples gave rise to 1000 representative states using \emph{K-means} clustering. We perform a grid search on a validation set from 0.001 to 1000 to identify the optimal kernel width. For the PPO method, we use 2 hidden layers, each consisting of 256 hidden units with \emph{tanh} nonlinear activation function at the first hidden layer. We set the default parameter values as follows: the discounting factor $\gamma = 0.99$, minibatch size $b = 128$, learning rate $lr=0.0005$ and clip parameter $\epsilon=0.3$. In addition, we use a mean standard deviation filter to normalize the state features. We train the deep MARL and DDMARL for 1 million iterations which on an average takes 48 hours on the GPU. The ensemble approach is trained for ~0.5 million iterations. 

\begin{figure}[!htb]
	\centering
	\begin{subfigure}{0.238\textwidth}
		\includegraphics[width=\textwidth]{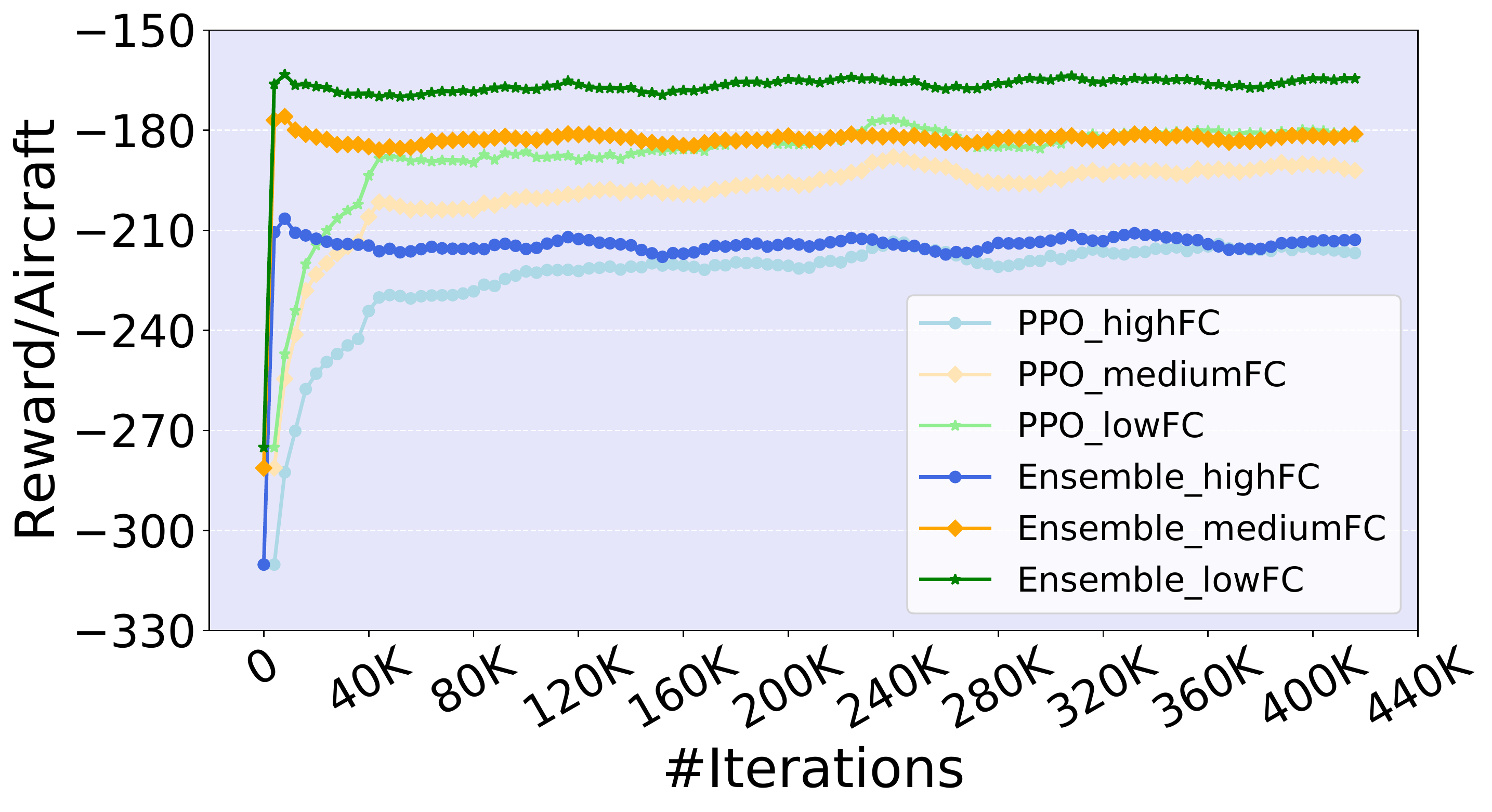} \caption{}
	\end{subfigure} 
	\begin{subfigure}{0.238\textwidth}
		\includegraphics[width=\textwidth]{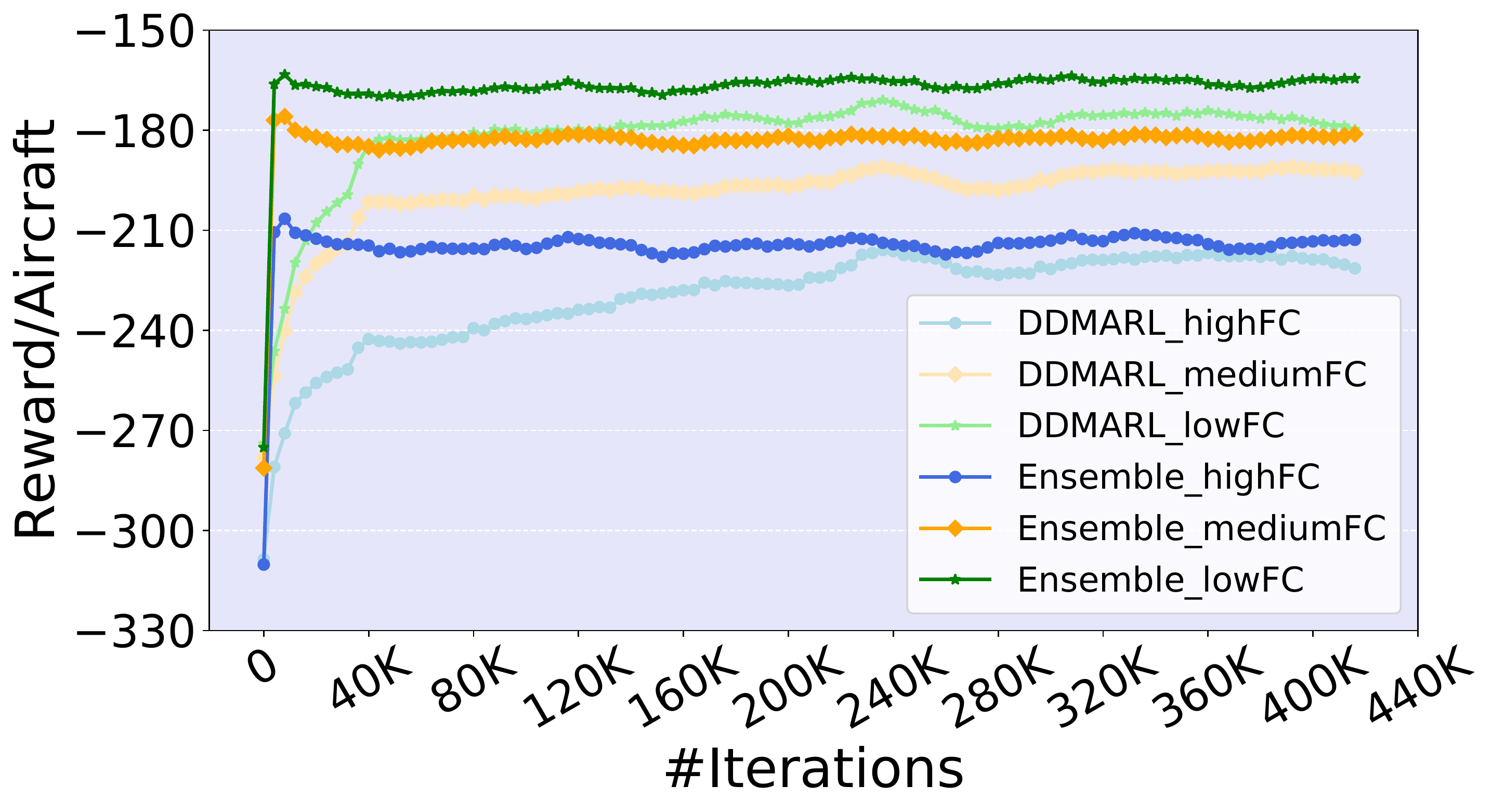} \caption{}
	\end{subfigure} 
	\caption{\small Training performance of (a) PPO Deep MARL vs. Deep Ensemble MARL; and (b) DDMARL vs. Deep Ensemble MARL.}
	\label{fig:drl_training}
\end{figure}

Figure \ref{fig:drl_training}(a) illustrates the training performance of PPO and our proposed deep Ensemble method for the three fuel cost settings. The Y-axis represents the expected reward per aircraft-agent and the X-axis denotes the iteration number. In all three settings, our proposed deep Ensemble method converges much faster than the PPO approach. We note that the Ensemble method does have the advantage that it starts with two pre-trained policies as its ``actions'' -- although learning to properly exploit both policies is still non-trivial.

For all the settings, the Ensemble learner parameters initially overfit to the initial training scenarios. This leads to the somewhat surprising early peak in 
the training performance  within first few interactions. Then, as model training  continues, the learner is exposed to other, different training scenarios which help to stabilize the learning process.  This initial overfitting effect can be easily confirmed experimentally by running the policies obtained at the early reward peak in the training phase to the test data and observing that those policies perform less well than the ones obtained after more extensive training, in spite of the slightly lower training phase reward of the latter.

A similar pattern can be seen with respect to the benchmark approach DDMARL, as illustrated in Figure \ref{fig:drl_training}(b). 
It should be noted that Figure \ref{fig:drl_training}(b) does not  provide a comparison of  computational times across the methods, because the Ensemble method is making use of two pre-trained policies. The computation time needed to train each constituent method is not shown in the figure.
The key findings from Figure \ref{fig:drl_training}(b) are two-fold: (a)
The slower convergence of the Ensemble method after the initial peak leads to a more robust solution than that of the initial peak and which offers better testing performance; and (b) The average  reward  after convergence of the ensemble method outperforms that of DDMARL. 

\subsection{Performance comparison on test data sets}

\begin{figure}[!htb]
	\centering
	\includegraphics[width=0.45\textwidth]{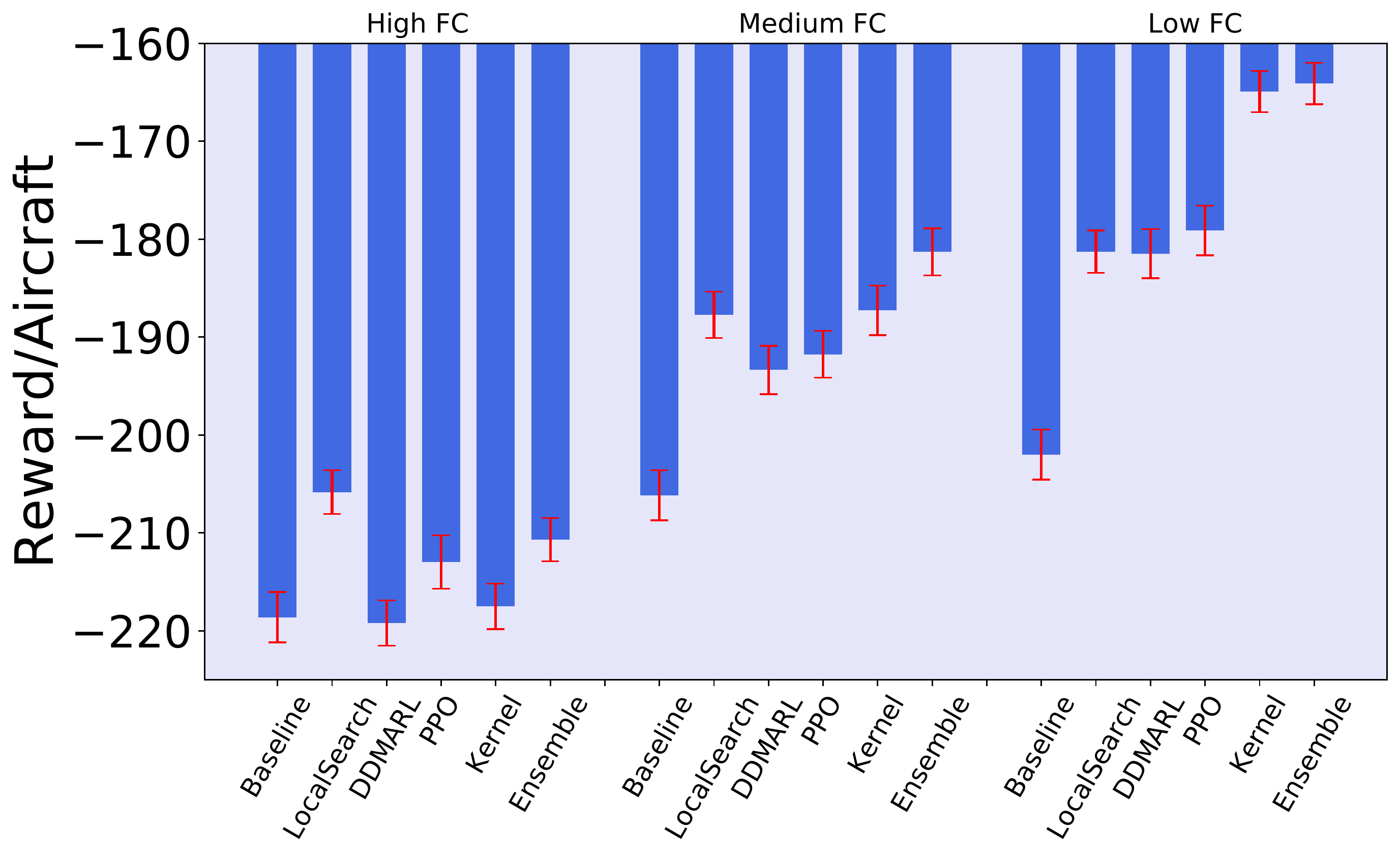}
	\caption{\small Performance comparison on 30 testing scenarios.}
	\label{fig:drl_testing}
\end{figure}

We evaluate the performance of our deep Ensemble method against the three benchmark approaches on 30 test scenarios. Figure \ref{fig:drl_testing} depicts the average reward per aircraft-agent for our proposed method as well as the component PPO and Kernel KBSF methods, for the 3 fuel cost settings. 
The red "I"-shaped indicators provide the standard error of the expected reward. 
In the case of a high fuel cost, the baseline schedule provides a near optimal solution as deviating from the scheduled speed incurs a significant fuel cost-related penalty. 
The reward associated with the baseline schedule is quite close to the solution of DDMARL.  By myopically improving the
baseline schedule, the local search method managed to
achieve a 5.5\% gain over the baseline.
Both PPO and our proposed Ensemble method provide around 3.6\% expected reward gain over the baseline.

The medium fuel cost scenario can be considered the most realistic of the three fuel cost settings.
In this setting, the proposed Ensemble method significantly outperforms the other approaches. On average, DD-MARL, local search, PPO and Kernel provide 6.2\%, 8.9\%, 7\% and 9.2\% gain in the expected reward over the baseline, respectively. The Ensemble method provides a 12.1\% improvement beyond the baseline. 

For low fuel costs, the Kernel approach is able to significantly outperform PPO and all other approaches. In this case, the deep Ensemble method is able to match the performance of the kernel-based approach.  The average performance gain of DD-MARL, local search, PPO, and Kernel over baseline are 10.3\%, 10.1\%, 11.3\%, 18.3\%, respectively, while our proposed Ensemble method beats the baseline by 18.8\%.

\begin{figure}[!htb]
	\centering
	\includegraphics[width=0.45\textwidth]{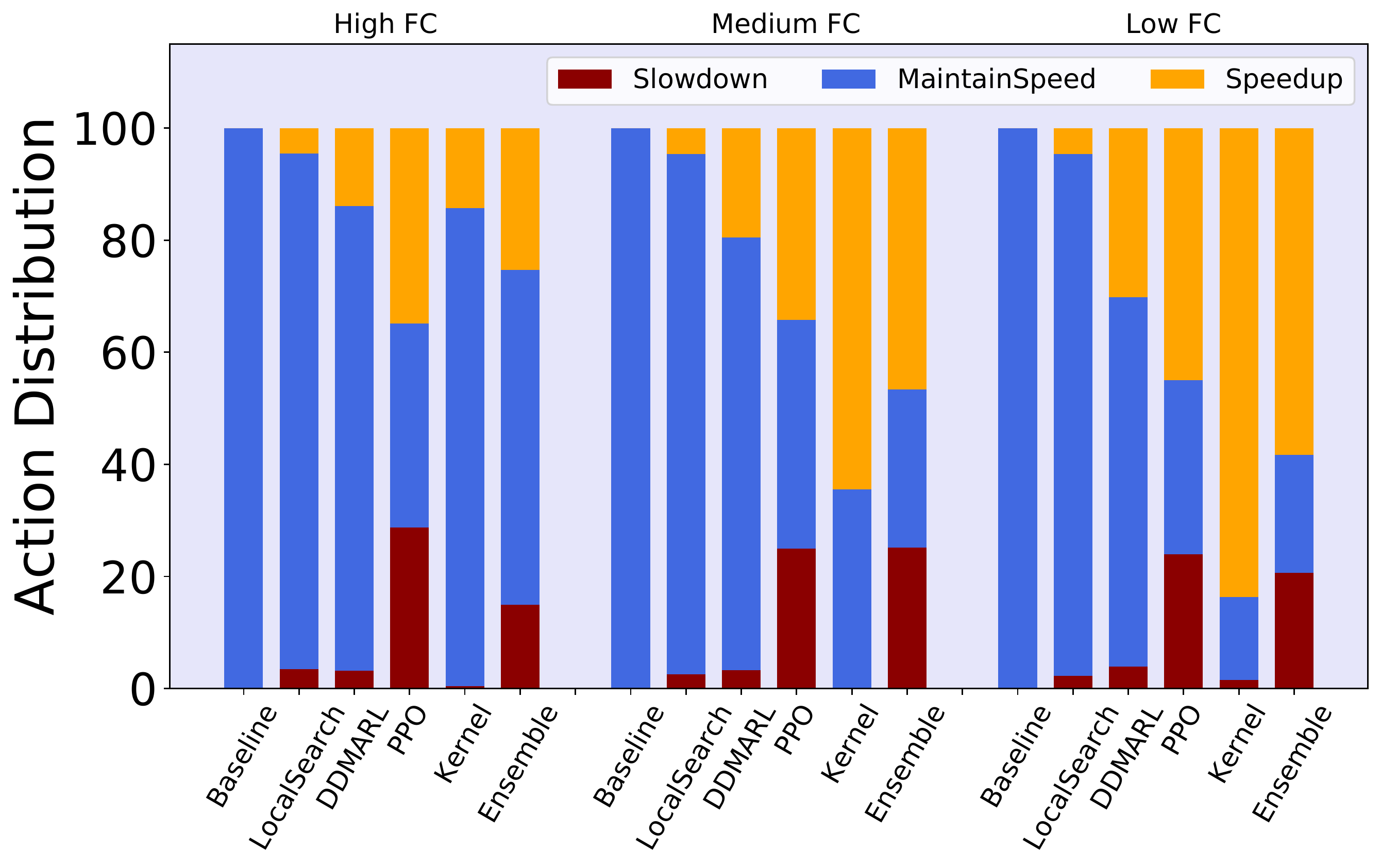}
	\caption{ Action distribution for different approaches on 30 testing scenarios.}
	\label{fig:action_distribution}
\end{figure}

Figure \ref{fig:action_distribution} illustrates the distribution of the actions taken for the three fuel cost settings. As expected, when the fuel cost is high, the action distribution of all approaches is skewed towards the baseline, since  there is little appetite for speed deviation in this case. 
In the low fuel cost scenario, the distribution is skewed towards speedup, which has the benefit of reducing delays and hence increasing reward. Notice, however, that the local search method is always skewed towards the baseline schedule and thus performs poorly especially in the low-cost case.
Our deep Ensemble method succeeds in leveraging the best of both worlds from the  fine-grained localized KBSF policies and the more global PPO policies; indeed the Ensemble method gains more  from the speedup action than does PPO, whilst not going to the extreme in proposing only speedup actions as is the case with the KBSF. 

The empirical results  demonstrate that the proposed Ensemble method can efficiently learn to arbitrate between the Kernel and deep MARL policies. The performance of the Ensemble method was further validated using the reward function of \citeauthor{brittain2019autonomous} \shortcite{brittain2019autonomous}, which takes into account only  a conflict cost. Notably, the Ensemble method outperforms theirs on their own reduced reward function by 17\%.

\section{Discussion}
We propose using reinforcement learning as a step towards the partial  automation  of real-time air traffic control for en-route aircraft management.
Specifically, we demonstrate that the complexity of the air traffic control problem can be concisely and accurately captured by learning an optimal deep ensemble with a multi-agent reinforcement learning (MARL) paradigm. The ensemble combines  a local kernel-based representation with a wider-reaching deep multi-agent reinforcement learning  model based on Proximal Policy Optimization (PPO). 
The results demonstrate the feasibility of using MARL for the problem of en-route air traffic control as well as the benefits of our proposed Ensemble method. 
This work marks a solid step into using reinforcement learning, and MARL in particular, for large-scale operational control. 

We see a number of avenues that merit further exploration to bring these approaches to the point where they can be operationally deployed.
The en-route aircraft management problem is possibly more amenable to MARL-based automation than the take off and landing problems. However, it will be important to extend in the action space with additional controls, including heading and altitude changes.
 In both of these extensions, we see that constrained RL has a role to play, to enable taking into account explicitly the complex business rules that govern those expanded control actions. 
 
\small
\bibliographystyle{named}
\bibliography{references}
\end{document}